%% file: main.tex
\pdfoutput=1
\documentclass[conference]{IEEEtran}
\usepackage{cite}
\usepackage{amsmath,amssymb,amsfonts}
\usepackage{algorithmic}
\usepackage{adjustbox}
\usepackage{graphicx}
\graphicspath{ {./figures/} }
\usepackage{textcomp}
\usepackage[table]{xcolor}
\usepackage[linesnumbered,ruled,vlined]{algorithm2e}
\usepackage{pdfcomment}
\usepackage{hyperref}
\usepackage{cleveref}
\usepackage{booktabs}
\usepackage{multicol}
\usepackage{multirow}
\usepackage{svg}
\usepackage[utf8]{inputenc}
\usepackage[T2A,T1]{fontenc}
\usepackage{stackengine}
\usepackage{multicol}
\usepackage{fancyhdr}
\def\ucr{\stackinset{c}{}{c}{-.2pt}{%
  \textcolor{white}{\sffamily\bfseries\small ?}}{%
  \rotatebox{45}{$\blacksquare$}}}

\begin{document}

\hypersetup{
    pdftitle={Bad Characters: Imperceptible NLP Attacks},
    pdfauthor={Nicholas Boucher, Ilia Shumailov, Ross Anderson, and Nicolas Papernot}
}

\newcommand{\utf}{\textbf{UTF-8}}
\newcommand{\unk}{\texttt{<unk>}\xspace}
\newcommand{\cyrillicx}{{\fontencoding{T2A}\selectfont х}\xspace}
\newcommand{\cyrillica}{{\fontencoding{T2A}\selectfont а}\xspace}
\newcommand{\cyrillicp}{{\fontencoding{T2A}\selectfont п}\xspace}
\newcommand{\cyrillicA}{{\fontencoding{T2A}\selectfont А}\xspace}
\newcommand{\zwnj}{{\fontfamily{cmr}\selectfont{‌}}}
\makeatletter
\newcount\my@repeat@count
\newcommand{\repeats}[2]{%
  \begingroup
  \my@repeat@count=\z@
  \@whilenum\my@repeat@count<#1\do{#2\advance\my@repeat@count\@ne}%
  \endgroup
}
\makeatother
\newcommand{\hundredzwnj}{\zwnj\zwnj\zwnj\zwnj\zwnj\zwnj\zwnj\zwnj\zwnj\zwnj\zwnj\zwnj\zwnj\zwnj\zwnj\zwnj\zwnj\zwnj\zwnj\zwnj\zwnj\zwnj\zwnj\zwnj\zwnj\zwnj\zwnj\zwnj\zwnj\zwnj\zwnj\zwnj\zwnj\zwnj\zwnj\zwnj\zwnj\zwnj\zwnj\zwnj\zwnj\zwnj\zwnj\zwnj\zwnj\zwnj\zwnj\zwnj\zwnj\zwnj\zwnj\zwnj\zwnj\zwnj\zwnj\zwnj\zwnj\zwnj\zwnj\zwnj\zwnj\zwnj\zwnj\zwnj\zwnj\zwnj\zwnj\zwnj\zwnj\zwnj\zwnj\zwnj\zwnj\zwnj\zwnj\zwnj\zwnj\zwnj\zwnj\zwnj\zwnj\zwnj\zwnj\zwnj\zwnj\zwnj\zwnj\zwnj\zwnj\zwnj\zwnj\zwnj\zwnj\zwnj\zwnj\zwnj\zwnj\zwnj\zwnj\zwnj}

\renewcommand{\algorithmiccomment}[1]{#1}

\makeatletter
\newcommand{\linebreakand}{%
  \end{@IEEEauthorhalign}
  \hfill\mbox{}\par
  \mbox{}\hfill\begin{@IEEEauthorhalign}
}
\makeatother

\fancypagestyle{firststyle}
{
   \fancyhf{}
   \chead{\textit{--------------------------------------- In 43rd IEEE Symposium on Security and Privacy ---------------------------------------}}
   \renewcommand{\headrulewidth}{0pt} 
}

\title{Bad\hundredzwnj\hundredzwnj\hundredzwnj\hundredzwnj\hundredzwnj\hundredzwnj\hundredzwnj\hundredzwnj\hundredzwnj\hundredzwnj{} Characters:\\Imperceptible NLP Attacks}

\author{\\\IEEEauthorblockN{Nicholas Boucher}
\IEEEauthorblockA{%
\textit{University of Cambridge}\\
\textit{Computer Science \& Technology}\\
nicholas.boucher@cl.cam.ac.uk}
\and
\\\IEEEauthorblockN{Ilia Shumailov}
\IEEEauthorblockA{%
\textit{University of Cambridge}\\
\textit{and Vector Institute}\\
ilia.shumailov@cl.cam.ac.uk}\\
\linebreakand
\IEEEauthorblockN{Ross Anderson}
\IEEEauthorblockA{%
\textit{University of Cambridge}\\
\textit{and University of Edinburgh}\\
ross.anderson@cl.cam.ac.uk}
\and
\IEEEauthorblockN{Nicolas Papernot}
\IEEEauthorblockA{%
\textit{University of Toronto}\\
\textit{and Vector Institute}\\
nicolas.papernot@utoronto.ca}
}

\maketitle

\thispagestyle{firststyle}

\input{sections/abstract}


\input{sections/intro}
\input{sections/motivation}
\input{sections/related}
\input{sections/background}
\input{sections/attacks}
\input{sections/evaluation}
\input{sections/discussion}
\input{sections/conclusion}

\onecolumn

\begin{multicols}{2}

\section*{Acknowledgment}

This work was supported by DARPA (through the GARD program), CIFAR (through a Canada CIFAR AI Chair), by NSERC (under the Discovery Program, and COHESA strategic research network), and by a gift from Intel. We also thank the Vector Institute's sponsors. Ilia Shumailov was supported with funds from Bosch-Forschungsstiftung im Stifterverband. We would also like to thank Adelin Travers for help with natural language annotation, Markus Kuhn and Zakhar Shumaylov for help with Unicode magic, and Darija Halatova for help with visualizations.

\bibliographystyle{IEEEtran}
\bibliography{sources}

\end{multicols}

\vfill

\appendix
\input{sections/appendix}

\end{document}

%% file: sections/abstract.tex
\begin{abstract}

Several years of research have shown that machine-learning systems are vulnerable to adversarial examples, both in theory and in practice. Until now, such attacks have primarily targeted visual models, exploiting the gap between human and machine perception. Although text-based models have also been attacked with adversarial examples, such attacks struggled to preserve semantic meaning and indistinguishability. In this paper, we explore a large class of adversarial examples that can be used to attack text-based models in a black-box setting without making any human-perceptible visual modification to inputs. We use encoding-specific perturbations that are imperceptible to the human eye to manipulate the outputs of a wide range of Natural Language Processing (NLP) systems from neural machine-translation pipelines to web search engines. We find that with a single imperceptible encoding injection -- representing one invisible character, homoglyph, reordering, or deletion -- an attacker can significantly reduce the performance of vulnerable models, and with three injections most models can be functionally broken. Our attacks work against currently-deployed commercial systems, including those produced by Microsoft and Google, in addition to open source models published by Facebook, IBM, and HuggingFace. This novel series of attacks presents a significant threat to many language processing systems: an attacker can affect systems in a targeted manner without any assumptions about the underlying model. We conclude that text-based NLP systems require careful input sanitization, just like conventional applications, and that given such systems are now being deployed rapidly at scale, the urgent attention of architects and operators is required.

\end{abstract}

%% file: sections/intro.tex
\section{Introduction}

\input{tables/examplestable}

Do {\fontfamily{Libertine}\selectfont x} and {\fontfamily{Libertine}\selectfont\cyrillicx} look the same to you? They may look identical to humans, but not to most natural-language processing systems. How many characters are in the string ``123\repeats{97}{\zwnj}''? If you guessed 100, you're correct. The first example contains the Latin character {\fontfamily{Libertine}\selectfont x} and the Cyrillic character h, which are typically rendered the same way. The second example contains 97 zero-width non-joiners\footnote{Unicode character U+200C} following the visible characters. Indeed, the title of this paper contains 1000 invisible characters imperceptible to human users.

Several years of research have demonstrated that machine-learning systems are vulnerable to adversarial examples, both theoretically and in practice~\cite{szegedy2013intriguing}. Such attacks initially targeted visual models used in image classification~\cite{goodfellow2015explaining}, though there has been recent interest in natural language processing and other applications. We present a broad class of powerful adversarial-example attacks on text-based models. These attacks apply input perturbations using invisible characters, control characters, and homoglyphs -- distinct characters with similar glyphs. These perturbations are imperceptible to human users, but the bytes used to encode them can change the output drastically. 

We have found that machine-learning models that process user-supplied text, such as neural machine-translation systems, are particularly vulnerable to this style of attack. Consider, for example, the market-leading service Google Translate~\cite{google_translate}. At the time of writing, entering the string ``paypal'' in the English to Russian model correctly outputs ``PayPal'', but replacing the Latin character a in the input with the Cyrillic character \cyrillica incorrectly outputs ``\cyrillicp\cyrillica\cyrillicp\cyrillica'' (``father'' in English). Model pipelines are agnostic of characters outside of their dictionary and replace them with \unk tokens; the software that calls them may however propagate unknown words from input to output. While that may help with general understanding of text, it opens a surprisingly large attack surface. 

Simple text-encoding attacks have been used occasionally in the past to get messages through spam filters. For example, there was a brief discussion in the SpamAssassin project in 2018 about how to deal with zero-width characters, which had been found in some sextortion scams~\cite{knight2018}. Although such tricks were known to engineers designing spam filters, they were not a primary concern. However, the rapid deployment of NLP systems in a large range of applications, from machine translation~\cite{Bengio2003ANP} through copyright enforcement~\cite{6970166} to hate speech filtering~\cite{schmidt-wiegand-2017-survey}, is suddenly creating a host of high-value targets that have capable motivated opponents.

The main contribution of this work is to explore and develop a class of imperceptible encoding-based attacks and to study their effect on the NLP systems that are now being deployed everywhere at scale. Our experiments show that many developers of such systems have been heedless of the risks; this is surprising given the long history of attacks on many varieties of systems that have exploited unsanitized inputs. We provide a set of examples of imperceptible attacks across various NLP tasks in~\Cref{tab:examples}. As we will later describe, these attacks take the form of invisible characters, homoglyphs, reorderings, and deletions injected via a genetic algorithm that maximizes a loss function defined for each NLP task.

Our findings present an attack vector that must be considered when designing any system processing natural language that may ingest text-based inputs with modern encodings, whether directly from an API or via document parsing. We then explore a series of defenses that can give some protection against this powerful set of attacks, such as discarding certain characters prior to tokenization, applying character mappings, and leveraging rendering and OCR for pre-processing. Defense is not entirely straightforward, though, as application requirements and resource constraints may prevent the use of specific defenses in certain circumstances.

This paper makes the following contributions:
\begin{itemize}
    \item We present a novel class of imperceptible perturbations for NLP models;
    \item We present four black-box variants of imperceptible attacks against both the integrity and availability of NLP models;
    \item We show that our imperceptible attacks degrade performance against task-appropriate benchmarks for eight models implementing machine translation, toxic content detection, textual entailment classification, named entity recognition, and sentiment analysis to near zero in untargeted attacks, succeed in most targeted attacks, and slow inference down by at least a factor of two in sponge example attacks;
    \item We evaluate our attacks extensively against both open source models and Machine Learning as a Service (MLaaS) offerings provided by Facebook, IBM, Microsoft, Google, and HuggingFace, finding that \textit{all} tested systems were vulnerable to three attack variants, and most were vulnerable to four;
    \item We present defenses against these attacks, and discuss why defense can be complex.
\end{itemize}

%% file: tables/examplestable.tex
\begin{table*}[t!]
\centering
\caption{Imperceptible Perturbations in Various NLP Tasks}
\label{tab:examples}
\begin{tabular}{llll}
\toprule
\textbf{Input Rendering}         & \textbf{Input Encoding}                         & \textbf{Task}                   & \textbf{Output} \\
\midrule
\rowcolor{gray!15}
Send money to account 1234       & Send money to account {\color{red}{U+202E}}4321 & Translation (EN$\rightarrow$FR) & \begin{tabular}[c]{@{}l@{}}Envoyer de l'argent au compte 4321 \\ \textit{(Send money to account 4321)}\end{tabular} \\
You are a coward and a fool.     & \begin{tabular}[c]{@{}l@{}}You ak{\color{red}{U+8}}re aq{\color{red}{U+8}} A{\color{red}{U+8}}coward and \\ a fov{\color{red}{U+8}}J{\color{red}{U+8}}ol.\end{tabular} & Toxic Content Detection         & \begin{tabular}[c]{@{}l@{}}8.2\% toxic \\ \textit{(96.8\% toxic unperturbed)}\end{tabular}  \\
\rowcolor{gray!15}
\begin{tabular}[c]{@{}l@{}}Oh, what a fool I feel! \\ / I am beyond proud.\end{tabular} & \begin{tabular}[c]{@{}l@{}}Oh, what a {\color{red}{U+200B}}fo{\color{red}{U+200B}}ol I{\color{red}{U+200B}} \\ {\color{red}{U+200B}}{\color{red}{U+200B}}feel! / I am beyond proud.\end{tabular} & Natural Language Inference      & \begin{tabular}[c]{@{}l@{}}0.3\% contradiction \\ \textit{(99.8\% contradiction unperturbed)}\end{tabular}   \\                                                                                               
\bottomrule
\end{tabular}
\end{table*}

%% file: sections/motivation.tex
\section{Motivation}
\label{sec:motivation}

Researchers have already experimented with adversarial attacks on NLP models~\cite{DBLP:journals/corr/PapernotMSH16,DBLP:journals/corr/abs-1711-02173,JiDeepWordBug18,ebrahimi-etal-2018-hotflip,DBLP:journals/corr/abs-1804-06059,zhao2018generating,alzantot-etal-2018-generating,li2019textbugger,michel-etal-2019-evaluation,ren-etal-2019-generating,shumailov2020sponge,zou2020reinforced}. However, up until now, such attacks were noticeable to human inspection and could be identified with relative ease. If the attacker inserts single-character spelling mistakes~\cite{DBLP:journals/corr/abs-1711-02173,JiDeepWordBug18,ebrahimi-etal-2018-hotflip,li2019textbugger}, they look out of place, while paraphrasing~\cite{DBLP:journals/corr/abs-1804-06059} often changes the meaning of a text enough to be noticeable. The attacks we discuss in this paper are the first class of attacks against modern NLP models that are imperceptible and do not distort semantic meaning. 

Our attacks can cause significant harm in practice. Consider two examples. First, consider a nation-state whose primary language is not spoken by the staff at a large social media company performing content moderation -- already a well-documented challenge~\cite{Frenkel2020}. If the government of this state wanted to make it difficult for moderators to block a campaign to incite violence against minorities, it could use imperceptible perturbations to stifle the efficacy of both machine-translation and toxic-content detection of inflammatory sentences.


Second, the ability to hide text in plain sight, by making it easy for humans to read but hard for machines to process, could be used by many bad actors to evade platform content filtering mechanisms and even impede law-enforcement and intelligence agencies. The same perturbations even prevent proper search-engine indexing, making malicious content hard to locate in the first place. We found that production search engines do not parse invisible characters and can be maliciously targeted with well-crafted queries. At the time of initial writing, Googling ``The meaning of life'' returned approximately 990 million results. Prior to responsible disclosure, searching for the visually identical string containing 250 invisible "zero width joiner" characters\footnote{Unicode character U+200D} returned exactly none.

%% file: sections/related.tex
\input{tables/taxonomy}

\section{Related work}
\label{sec:related}

\subsection{Adversarial Examples}

Machine-learning techniques are vulnerable to many large classes of attack~\cite{tabassi2019taxonomy}, with one major class being adversarial examples. These are inputs to models which, during inference, cause the model to output an incorrect result~\cite{szegedy2013intriguing}. In a white-box environment -- where the adversary knows the model -- such examples can be found using a number of gradient-based methods which typically aim to maximize the loss function under a series of constraints~\cite{szegedy2013intriguing,goodfellow2015explaining,madry2019deep}. In the black-box setting, where the model is unknown, the adversary can transfer adversarial examples from another model~\cite{papernot2017practical}, or approximate gradients by observing output labels and, in some settings, confidence~\cite{chen2017zoo}.

Training data can also be poisoned to manipulate the accuracy of the model for specific inputs~\cite{nelson2008exploiting,jagielski2018manipulating}. Bitwise errors can be introduced during inference to reduce the model's performance~\cite{hong2019terminalbrain}. Inputs can also be chosen to maximize the time or energy a model takes during inference~\cite{shumailov2020sponge}, or to expose confidential training data via inference techniques~\cite{choo2020labelonly}. In other words, adversarial algorithms can affect the \textit{integrity}, \textit{availability} and \textit{confidentiality} of machine-learning systems~\cite{biggio2018wild,papernot2016towards,shumailov2020sponge}.

\subsection{NLP Models}
Natural language processing (NLP) systems are designed to process human language. Machine translation was proposed as early as 1949~\cite{weaver_translation_1949} and has become a key sub-field of NLP. Early approaches to machine translation tended to be rule-based, using expert knowledge from human linguists, but statistical methods became more prominent as the field matured~\cite{dorr_survey_1998}, eventually yielding to neural networks~\cite{Bengio2003ANP}, then recurrent neural networks (RNNs) because of their ability to reference past context~\cite{kalchbrenner-blunsom-2013-recurrent}. The current state of the art is the Transformer model, which provides the benefits of RNNs and CNNs in a traditional network via the use of an attention mechanism~\cite{vaswani2017attention}.

Transformers are a form of encoder-decoder model~\cite{NIPS2014_a14ac55a,DBLP:journals/corr/ChoMGBSB14} that map sequences to sequences. Each source language has an encoder that converts the input into a learned interlingua, an intermediate representation which is then decoded into the target language using a model associated with that language.

Regardless of the details of the model used for translation, natural language must be encoded in a manner that can be used as its input. The simplest encoding is a dictionary that maps words to numerical representations, but this fails to encode previously unseen words and thus suffers from limited vocabulary. N-gram encodings can increase performance, but increase the dictionary size exponentially while failing to solve the unseen-word problem. A common strategy is to decompose words into sub-word segments prior to encoding, as this enables the encoding and translation of previously unseen words in many circumstances~\cite{DBLP:journals/corr/SennrichHB15}.

\subsection{Adversarial NLP}

Early adversarial ML research focused on image classification~\cite{biggio2013evasion,goodfellow2015explaining}, and the search for adversarial examples in NLP systems began later, targeting sequence models~\cite{DBLP:journals/corr/PapernotMSH16}. Adversarial examples are inherently harder to craft due to the discrete nature of natural language. Unlike images in which pixel values can be adjusted in a near-continuous and virtually imperceptible fashion to maximize loss functions, perturbations to natural language are more visible and involve the manipulation of more discrete tokens.

More generally, source language perturbations that will provide effective adversarial samples against human users need to account for semantic similarity~\cite{michel-etal-2019-evaluation}. Researchers have proposed using word-based input swaps with synonyms~\cite{ren-etal-2019-generating} or character-based swaps with semantic constraints~\cite{ebrahimi-etal-2018-hotflip}. These methods aim to constrain the perturbations to a set of transformations that a human is less likely to notice. Both neural machine-translation~\cite{DBLP:journals/corr/abs-1711-02173} and text classification~\cite{JiDeepWordBug18, li2019textbugger} models generally perform poorly on noisy inputs such as misspellings, but such perturbations create clear visual artifacts that are easier for humans to notice.

Using different paraphrases of the same meaning, rather than one-to-one synonyms, may give more leeway. Paraphrase sets can be generated by comparing machine back-translations of large corpora of text~\cite{wieting-etal-2017-learning}, and used to systematically generate adversarial examples for machine-translation systems~\cite{DBLP:journals/corr/abs-1804-06059}. One can also search for neighbors of the input sentence in an embedded space~\cite{zhao2018generating}; these examples often result in low-performance translations, making them candidates for adversarial examples. BLEU score is commonly used for assessing the quality of machine translations~\cite{papineni-etal-2002-bleu}, and therefore also for assessing related attacks. Although paraphrasing can indeed help preserve semantics, humans often notice that the results look odd. Our attacks on the other hand do not introduce any visible perturbations, use fewer substitutions, and preserve semantic meaning perfectly. 

Genetic algorithms have been used to find adversarial perturbations against inputs to sentiment analysis systems, presenting an attack viable in the black-box setting without access to gradients~\cite{alzantot-etal-2018-generating}. Reinforcement learning can be used to efficiently generate adversarial examples for translation models~\cite{zou2020reinforced}. There have even been efforts to combine academic NLP adversarial techniques into easily consumable toolkits available online~\cite{morris2020textattack}, making these attacks relatively easy to use. Unlike the techniques described in this paper, though, all existing NLP adversarial example techniques result in human-perceptible visual artifacts within inputs.

Michel et al. also propose that unknown tokens \unk, which are used to encode text sequences not recognized by the natural language encoder in NLP settings, can be leveraged to make compelling source language perturbations due to the flexibility of the characters which encode to \unk~\cite{michel-etal-2019-evaluation}. However, all methods proposed so far for generating \unk use visible characters.

We present a taxonomy of adversarial NLP attacks in \Cref{tab:taxonomy}.

\subsection{Unicode}

Unicode is a character set designed to standardize the electronic representation of text~\cite{unicode_2020}. As of the time of writing, it can represent 143,859 characters across many different languages and symbol groups. Characters as diverse as Latin letters, traditional Chinese characters, mathematical notation, and emojis can all be represented in Unicode. It maps each character to a code point, or numerical representation.

These numerical code points, often denoted with the prefix \texttt{U+}, can be encoded in a variety of ways, although UTF-8 is the most common. This is a variable-length encoding scheme that represents code points as 1-4 bytes. 

A font is a collection of glyphs that describe how code points should be rendered. Most computers support many different fonts. It is not required that fonts have a glyph for every code point, and code points without corresponding glyphs are typically rendered as an `unknown' placeholder character.


\subsection{Unicode Security}
\label{sec:unicode_security}

As it has to support a globally broad set of languages, the Unicode specification is quite complex. This complexity can lead to security issues, as detailed in the Unicode Consortium's technical report on Unicode security considerations~\cite{unicode_security_2014}.

One primary security consideration in the Unicode specification is the multitude of ways to encode homoglyphs, which are unique characters that share the same or nearly the same glyph. This problem is not unique to Unicode; for example, in the ASCII range, the rendering of the lowercase Latin `l'\footnote{ASCII value \texttt{0x6C}} is often  nearly identical to the uppercase Latin `I'\footnote{ASCII value \texttt{0x49}}. In some fonts, character sequences can act as pseudo-homoglyphs, such as the sequences `rn' and `m' in most sans serif fonts.

Such visual tricks provide a tool in the arsenal of cyber scammers~\cite{SimpsonMC20apwg}. The earliest example we found is that of \textit{paypaI.com} (notice the last domain name character is an uppercase `I'), which was used in July 2000 to trick users into disclosing passwords for \textit{paypal.com}~\cite{sullivan_paypal_2000}. Indeed, significant attention has since been given to homoglyphs in URLs~\cite{10.1145/503124.503156,10.5555/1267359.1267383,noauthor_capec-632_2015,10.1145/3355369.3355587}. Some browsers attempt to remedy this ambiguity by rendering all URL characters in their lowercase form upon navigation, and the IETF set a standard to resolve ambiguities between non-ASCII characters that are homoglyphs with ASCII characters. This standard, called Punycode, resolves non-ASCII URLs to an encoding restricted to the ASCII range. For example, most browsers will automatically re-render the URL \textit{p\cyrillica{}yp\cyrillica{}l.com} (which uses the Cyrillic \cyrillica\footnote{Unicode character U+0430}) to its Punycode equivalent \textit{xn--pypl-53dc.com} to highlight a potentially dangerous ambiguity. However, Punycode can introduce new opportunities for deception. For example, the URL \textit{xn--google.com} decodes to four semantically meaningless traditional Chinese characters. Furthermore, Punycode does not solve cross-script homoglyph encoding vulnerabilities outside of URLs. For example, homoglyphs have in the past caused security vulnerabilities in various non-URL areas such as certificate common names.

Homoglyphs have also been proposed for information hiding, such as encoding information via sequences of different whitespace characters~\cite{POR20121075}.

Unicode attacks can also exploit character ordering. Some character sets (such as Hebrew and Arabic) naturally display in right-to-left order. The possibility of intermixing left-to-right and right-to-left text, as when an English phrase is quoted in an Arabic newspaper, necessitates a system for managing character order with mixed character sets. For Unicode, this is the Bidirectional (Bidi) Algorithm~\cite{unicode_bidi_2020}. Unicode specifies a variety of control characters that allow a document creator to fine-tune character ordering, including Bidi override characters that allow complete control over display order. The net effect is that an adversary can force characters to render in a different order than they are encoded, thus permitting the same visual rendering to be represented by a variety of different encoded sequences. Historically, Bidi overrides have been used by scammers to change the appearance of file extensions, thus enabling stealthy dissemination of malware~\cite{brian_krebs_right--left_2011}.

Lastly, an entire class of vulnerabilities stems from bugs in Unicode implementations. These have historically been used to generate a range of interesting exploits about which it is difficult to generalize. While the Unicode Consortium does publish a set of software components for Unicode support~\cite{icu_2021}, many operating systems, platforms, and other software ecosystems have different implementations. For example, GNOME produces Pango~\cite{pango}, Apple produces Core Text~\cite{apple_coretext}, while Microsoft produces a Unicode implementation for Windows~\cite{microsoft_unicode}. 

In what follows, we will mostly disregard bugs and focus on attacks that exploit correct implementations of the Unicode standard. We instead exploit the gap between visualization and NLP pipelines.

%% file: tables/taxonomy.tex
\begin{table*}[t]
\centering
\caption{Taxonomy of Adversarial NLP attacks in academic literature.}
\label{tab:taxonomy}
\begin{tabular}{lccc|cc|c}
\toprule
\multirow{2}{*}{\textbf{Attack}}                                                      & \multicolumn{3}{c|}{\textbf{Features}}                                                                     & \multicolumn{2}{c|}{\textbf{Integrity}}                                          & \textbf{Availability}            \\
                                                                                      & Imperceptible                     & Semantic Similarity                     & Blackbox                     & Classification                           & Translation                           & DoS                              \\ \midrule
\rowcolor{gray!15}
RNN Adversarial Sequences~\cite{DBLP:journals/corr/PapernotMSH16}                     &                                   &                                         &                              & \checkmark                               &                                       &                                  \\
Synthetic and Natural Noise~\cite{DBLP:journals/corr/abs-1711-02173}                  &                                   &                                         & \checkmark                   &                                          & \checkmark                            &                                  \\
\rowcolor{gray!15}
DeepWordBug~\cite{JiDeepWordBug18}                                                    &                                   &                                         & \checkmark                   & \checkmark                               &                                       &                                  \\
HotFlip~\cite{ebrahimi-etal-2018-hotflip}                                             &                                   &                                         &                              & \checkmark                               &                                       &                                  \\
\rowcolor{gray!15}
Syntactically Controlled Paraphrase~\cite{DBLP:journals/corr/abs-1804-06059}          &                                   & \checkmark                              & \checkmark                   & \checkmark                               &                                       &                                  \\
Natural Adversarial Examples~\cite{zhao2018generating}                                &                                   &                                         & \checkmark                   & \checkmark                               & \checkmark                            &                                  \\
\rowcolor{gray!15}
Natural Language Adversarial Examples~\cite{alzantot-etal-2018-generating,jia2019certified}            &                                   & \checkmark                              & \checkmark                   & \checkmark                               &                                       &                                  \\
TextBugger~\cite{li2019textbugger}                                                    &                                   &                                         & \checkmark                   & \checkmark                               &                                       &                                  \\
\rowcolor{gray!15}
seq2seq Adversarial Perturbations~\cite{michel-etal-2019-evaluation}                  &                                   & \checkmark                              &                              &                                          & \checkmark                            &                                  \\
Probability Weighted Word Saliency~\cite{ren-etal-2019-generating}                    &                                   & \checkmark                              &                              & \checkmark                               &                                       &                                  \\
\rowcolor{gray!15}
Sponge Examples~\cite{shumailov2020sponge}                                            &                                   &                                         & \checkmark                   &                                &                             & \checkmark                       \\
Reinforced Generation~\cite{zou2020reinforced}                                        &                                   & \checkmark                              & \checkmark                   &                                          & \checkmark                            &                                  \\
\rowcolor{gray!15}
\textbf{Imperceptible Perturbations}                                                  & \checkmark                        & \checkmark                              & \checkmark                   & \checkmark                               & \checkmark                            & \checkmark                       \\ \bottomrule
\end{tabular}
\end{table*}

%% file: sections/background.tex
\section{Background}

\subsection{Attack Taxonomy}
In this paper, we explore the class of imperceptible attacks based on Unicode and other encoding conventions which are generally applicable to text-based NLP models. We see each attack as a form of adversarial example whereby imperceptible perturbations are applied to fixed inputs into existing text-based NLP models.

We define these \textit{imperceptible perturbations} as modifications to the encoding of a string of text which result in either:
\begin{itemize}
    \item No visual modification to the string's rendering by a standards-compliant rendering engine compared to the unperturbed input, or
    \item Visual modifications sufficiently subtle to go unnoticed by the average human reader using common fonts. 
\end{itemize}

For the latter case, it is alternatively possible to replace human imperceptibility as indistinguishability by a computer vision model between images of the renderings of two strings, or a maximum pixel-wise difference between such rendering.

We consider four different classes of imperceptible attack against NLP models:

\begin{enumerate}
    \item \textbf{Invisible Characters}: Valid characters which by design do not render to a visible glyph are used to perturb the input to a model.
    \item \textbf{Homoglyphs}: Unique characters which render to the same or visually similar glyphs are used to perturb the input to a model.
    \item \textbf{Reorderings}: Directionality control characters are used to override the default rendering order of glyphs, allowing reordering of the encoded bytes used as input to a model.
    \item \textbf{Deletions}: Deletion control characters, such as the backspace, are injected into a string to remove injected characters from its visual rendering to perturb the input to a model.
\end{enumerate}

These imperceptible text-based attacks on NLP models represent an abstract class of attacks independent of different text-encoding standards and implementations. For the purpose of concrete examples and experimental results, we will assume the near-ubiquitous Unicode encoding standard, but we believe our results to be generalizable to any encoding standard with a sufficiently large character and control-sequence set.

Further classes of text-based attacks exist, as detailed in \Cref{tab:examples}, but all other attack classes produce visual artifacts.

The imperceptible text-based attacks described in this paper can be used against a broad range of NLP models. As we explain later, imperceptible perturbations can manipulate machine translation, break toxic content classifiers, degrade search engine querying and indexing, and generate sponge examples~\cite{shumailov2020sponge} for denial-of-service (DoS) attacks, among other possibilities.

\subsection{NLP Pipeline}

Modern NLP pipelines have evolved through decades of research to include a large number of performance optimizations. Text-based inputs undergo a number of pre-processing steps before model inference. Typically a \textit{tokenizer} is first applied to separate words and punctuation in a task-meaningful way, an example being the Moses tokenizer~\cite{koehn2007moses} used in the Fairseq models evaluated later in this paper. Tokenized words are then encoded. Early models used dictionaries to map tokens to encoded embeddings, and tokens not seen during training were replaced with a special \unk embedding. Many modern models now apply Byte Pair Encoding (BPE) or the WordPiece algorithm~\cite{wu2016googles} before dictionary lookups. BPE, a common data compression technique, and WordPiece both identify common subwords in tokens. This often results in increased performance, as it allows the model to capture additional knowledge about language semantics from morphemes \cite{sennrich2015subword}. Both of these pre-processing methodologies are commonly used in deployed NLP models, including all five open source models published by Facebook, IBM, and HuggingFace evaluated in this paper.

Modern NLP pipelines process text in a very different manner than text-rendering systems, even when dealing with the same input. While the NLP system is dealing with the semantics of human language, the rendering engine is dealing with a large, rich set of different control characters. This structural difference between what models see and what humans see is what we exploit in our attacks.

\subsection{Attack Methodology}
\label{sec:attack_method}

We approach the generation of adversarial samples as an optimization problem. Assume an NLP function $f(\mathbf{x})=\mathbf{y}: X\rightarrow Y$ mapping textual input $\mathbf{x}$ to $\mathbf{y}$. Depending on the task, $Y$ is either a sequence of characters, words, or hot-encoded categories. For example, translation tasks such as WMT assume $Y$ to be a sequence of characters, whereas categorization tasks such as MNLI assume $Y$ to be one of three categories.  Furthermore, we assume a strong black-box threat model where adversaries have access to model output but cannot observe the internals. This makes our attack realistic: we later show it can be mounted on existing commercial ML services. In this threat model, an adversary's goal is to imperceptibly manipulate $f$ using a perturbation function $p$.

These manipulations fall into two categories:

\begin{itemize}
    \item \textbf{Integrity Attack}: The adversary aims to find $p$ such that $f(p(\mathbf{x})) \neq f(\mathbf{x})$. For a targeted attack, the adversary further constrains $p$ such that the perturbed output matches a fixed target $\mathbf{t}$: $f(p(\mathbf{x})) = \mathbf{t}$.
    \item \textbf{Availability Attack}: The adversary aims to find $p$ such that $\mathit{time}(f(p(\mathbf{x}))) > \mathit{time}(f(\mathbf{x}))$, where \textit{time}  measures the inference runtime of $f$.  
\end{itemize}

We also define a constraint on the perturbation function $p$:
\begin{itemize}
    \item \textbf{Budget}: A budget $b$ such that $\mathit{dist}(\mathbf{x}, p(\mathbf{x})) \leq b$. The function \textit{dist} may refer to any distance metric.
\end{itemize}

We define the attack as optimizing a set of operations over the input text, where each operation corresponds to the injection of one short sequence of Unicode characters to perform a single imperceptible perturbation of the chosen class. The length of the injected sequence is dependent upon the class chosen and attack implementation; in our evaluation we use one character injections for invisible characters and homoglyphs, two characters for deletions, and ten characters for reorderings, as later described. We select a gradient-free optimization method -- differential evolution~\cite{storn_differential_1997} -- to enable this attack to work in the black-box setting without having to recover approximated gradients. This approach randomly initializes a set of candidates and evolves them over many iterations, ultimately selecting the best-performing traits. 
 
The attack algorithm is shown in~\Cref{alg:attackalg}. It takes as parameters input text $x$ and attack $\mathcal{A}$, representing either an invisible character, homoglyph, reordering, or deletion attack. $\mathcal{A}$ is a function which applies its attack according to the parameters passed to it encoding the location and degree of the perturbation, bounded by $\mathcal{B}_\mathcal{A}$ according to budget $\beta$. It also takes a model $\mathcal{T}$ implementing an NLP task, and optionally a target output $y$ if performing a targeted attack. Finally, it expects parameters representing a population size $s$, number of evolutions $m$, differential weight $F$, and crossover probability $CR$, which are all standard parameters of differential evolution optimization~\cite{storn_differential_1997}. In summary, the attack algorithm defines an objective function $\mathcal{F}(\cdot)$, which seeks to either maximize perturbed model output Levenshtein distance from its unperturbed output, minimize model output Levenshtein distance to a target value, or maximize model inference time. This objective function is then optimized using differential evolution, a common gradient-free genetic optimization. Finally, the perturbed text optimizing the objective function $\mathcal{F}(\cdot)$ is returned.

\subsection{Invisible Characters}

\input{figures/infos/invis_char}

\input{tables/attack_algo}

Invisible characters are encoded characters that render to the absence of a glyph and take up no space in the resulting rendering. Invisible characters are typically not font-specific, but follow from the specification of an encoding format. An example in Unicode is the zero-width space character\footnote{Unicode character U+200B} (ZWSP). An example of an attack using invisible characters is shown in~\Cref{fig:invis_attack}.

It is important to note that characters lacking a glyph definition in a specific font are not typically treated as invisible characters. Due to the number of characters in Unicode and other large specifications, fonts will often omit glyph definitions for rare characters. For example, Unicode supports characters from the ancient Mycenaean script Linear B, but these glyph definitions are unlikely to appear in fonts targeting modern languages such as English. However, most text-rendering systems reserve a special character, often $\square$ or \ucr, for valid Unicode encodings with no corresponding glyph. These characters are therefore visible in rendered text.

In practice, though, invisible characters are font-specific. Even though some characters are designed to have a non-glyph rendering, the details are up to the font designer. They might, for example, render all traditionally invisible characters by printing the corresponding Unicode code point as a base 10 numeral. Yet a small number of fonts dominate the modern world of computing, and fonts in common use are likely to respect the spirit of the Unicode specification. For the purposes of this paper, we will determine character visibility using GNU's Unifont~\cite{unifont} glyphs. Unifont was chosen because of its relatively robust coverage of the current Unicode standard, its distribution with common operating systems, and its visual similarity to other common fonts.

Although invisible characters do not produce a rendered glyph, they nevertheless represent a valid encoded character. Text-based NLP models operate over encoded bytes as inputs, so these characters will be ``seen'' by a text-based model even if they are not rendered to anything perceptible by a human user. We found that these bytes alter model output. When injected arbitrarily into a model's input, they typically degrade the performance both in terms of accuracy and runtime. When injected in a targeted fashion, they can be used to modify the output in a desired way, and may coherently change the meaning of the output across many NLP tasks.

\subsection{Homoglyphs}
\label{sec:homoglyphs}

\input{figures/infos/homoglyphs}

Homoglyphs are characters that render to the same glyph or to a visually similar glyph. This often occurs when portions of the same written script are used across different language families. For example, consider the Latin letter `A' used in English. The very similar character `\cyrillicA' is used in the Cyrillic alphabet. Within the Unicode specification these are distinct characters, although they are typically rendered as homoglyphs.

An example of an attack using homoglyphs is shown in~\Cref{fig:homoglyphs_attack}. Like invisible characters, homoglyphs are font-specific. Even if the underlying linguistic system denotes two characters in the same way, fonts are not required to respect this. That said, there are well-known homoglyphs in the most common fonts used in everyday computing.

The Unicode Consortium publishes two supporting documents with the Unicode Security Mechanisms technical report~\cite{unicode_security_2020} to draw attention to similarly rendered characters. The first defines a mapping of characters that are intended to be homoglyphs within the Unicode specification and should therefore map to the same glyph in font implementations~\cite{unicode_intentionals}. The second document~\cite{unicode_confusables} defines a set of characters that are likely to be visually confused, even if they are not rendered with precisely the same glyph.

For the experiments in this paper, we use the Unicode technical reports to define homoglpyh mappings. We also note that homoglyphs, particularly for specific less common fonts, can be identified using an unsupervised clustering algorithm against vectors representing rendered glyphs. To illustrate this, we used a VGG16 convolution neural network~\cite{Simonyan15} to transform all glyphs in the Unifont font into vectorized embeddings and performed various clustering operations. \Cref{fig:homoglyph-pca} visualizes mappings provided by the Unicode technical reports as a dimensionality-reduced character cluster plot. We find that the results of well-tuned unsupervised clustering algorithms produce similar results, but have chosen to use the official Unicode mappings in this paper for reproducibility.

\subsection{Reorderings}

\input{figures/infos/reordering}

\input{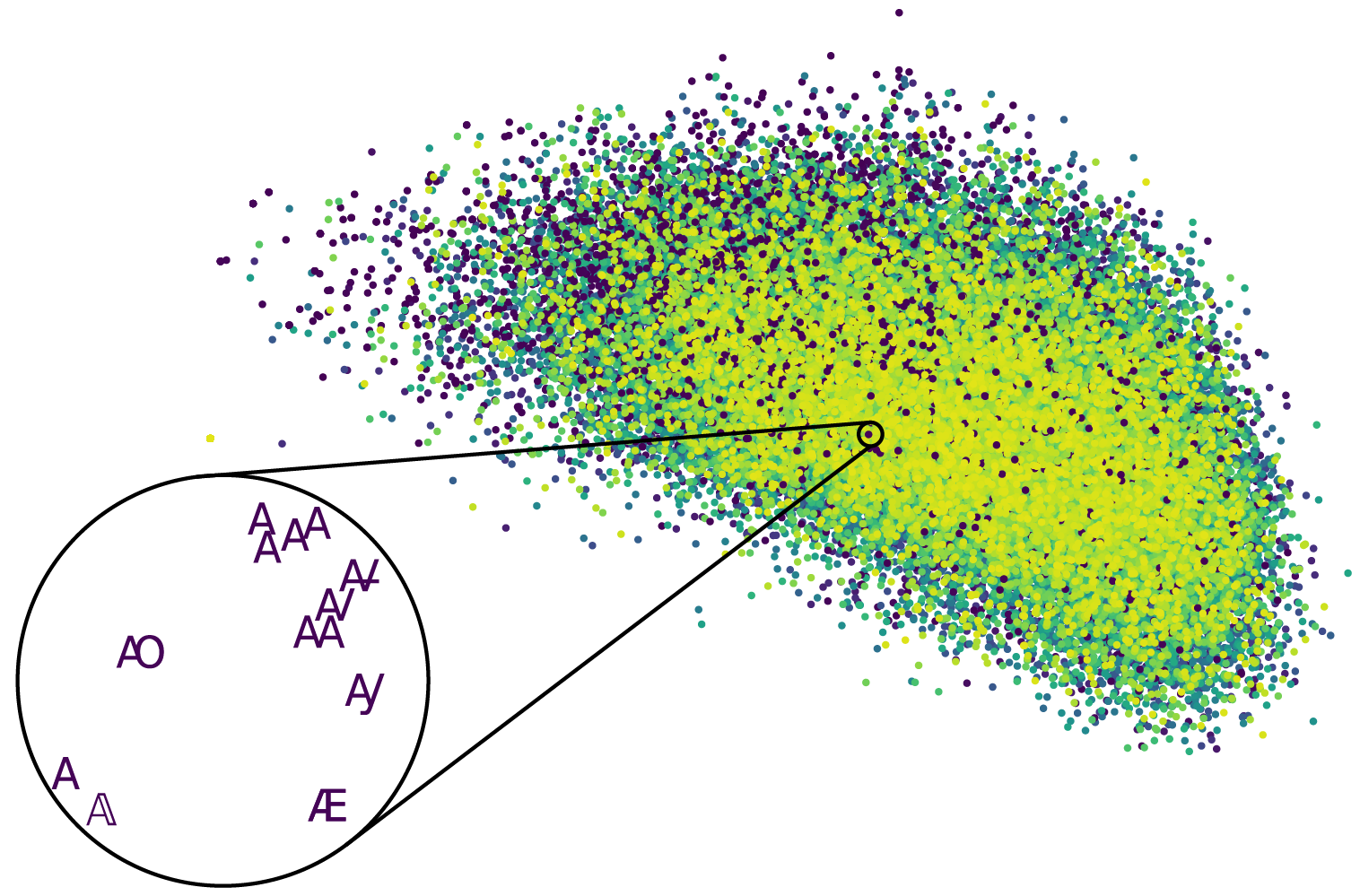}

The Unicode specification supports characters from languages that read in both the left-to-right and right-to-left directions. This becomes nontrivial to manage when such scripts are mixed. The Unicode specification defines the Bidirectional (Bidi) Algorithm~\cite{unicode_bidi_2020} to support standard rendering behavior for mixed-script documents. However, the specification also allows the Bidi Algorithm to be overridden using invisible direction-override control characters, which allow near-arbitrary rendering for a fixed encoded ordering.

An example of an attack using reorderings is shown in~\Cref{fig:reorder_attack}. In an adversarial setting, Bidi control characters allow the encoded ordering of characters to be shuffled without affecting character rendering thus making them a form of imperceptible perturbation.

Unlike invisible character and homoglyph attacks, the class of reordering attacks is font-independent and relies only on the implementation of the Unicode Bidi Algorithm. Bidi algorithm implementations sometimes differ in how they handle specific control sequences, meaning that some attacks may be platform or application specific in practice, but most mature Unicode rendering systems behave similarly. Appendix \Cref{alg:reorderalg} defines an algorithm for generating $2^{n-1}$ unique reorderings for strings of length $n$ using nested Bidi control characters. At the time of writing, it has been tested to work against the Unicode implementation in Chromium~\cite{google_chromium}.


\subsection{Deletions}

\input{figures/infos/backspace}

A small number of control characters in Unicode can cause neighbouring text to be removed. The simplest examples are the backspace (BS) and delete (DEL) characters. There is also the carriage return (CR) which causes the text-rendering algorithm to return to the beginning of the line and overwrite its contents. For example, encoded text which represents ``Hello \textbf{\texttt{CR}}Goodbye World'' will be rendered as ``Goodbye World''.

An example of an attack using deletions is shown in~\Cref{fig:backspace_attack}. Deletion attacks are font-independent, as Unicode does not allow glyph specification for the basic control characters inherited from ASCII including BS, DEL, and CR. In general, deletion attacks are also platform independent as there is not significant variance in Unicode deletion implementations. However, these attacks can be harder to exploit in practice because most systems do not copy deleted text to the clipboard. As such, an attack using deletion perturbations generally requires an adversary to submit encoded Unicode bytes directly into a model, rather than relying on a victim's copy+paste functionality.

%% file: figures/infos/invis_char.tex
\begin{figure}[t]
    \centering
    \includegraphics[width=\linewidth]{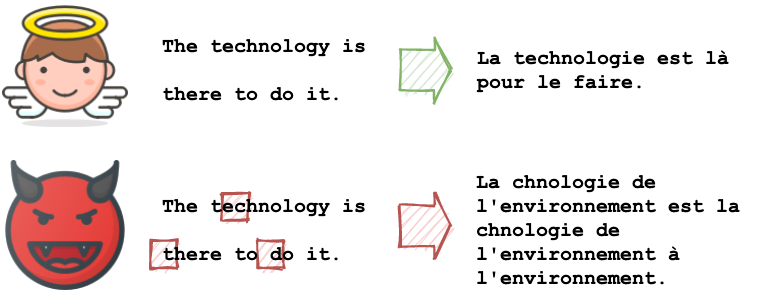}
    \caption{Attack using invisible characters. Example machine translation input is on the left with model output on the right. Invisible characters are denoted by red boxes, such as between the `e' and `c'.}
    \label{fig:invis_attack}
\end{figure}

%% file: tables/attack_algo.tex
\begin{algorithm}[t]
   \caption{Imperceptible perturbations adversarial example via differential evolution.}
   \label{alg:attackalg}
\begin{algorithmic}
    \STATE {\bfseries Input:} text $\mathbf{x}$, attack $\mathcal{A}$ with input bounds distribution $\mathcal{B}_\mathcal{A}$, NLP task $\mathcal{T}$, target $\mathbf{y}$, perturbation budget $\beta$, population size $s$, evolution iterations $m$, differential weight $F \in [0,2]$, crossover probability $\mathit{CR} \in [0,1]$
    \STATE \textbf{Result:} Adversarial example visually identical to \textbf{x} against task $\mathcal{T}$ using attack $\mathcal{A}$
    \STATE
    
   \STATE{\begin{tabular}{@{}r@{}l@{}}
    Randomly initialize population $\mathbf{P} := \{\mathbf{p_0}, \dots, \mathbf{p_s}\}$&, \\
    where $\mathbf{p_n} \sim \mathcal{B}_\mathcal{A}(\mathbf{x})$&
    \end{tabular}}
    \IF{availability attack}
        \STATE{$\mathcal{F}(\cdot) = \text{execution\_time}(\mathcal{T}(\mathcal{A}(\mathbf{x}, \cdot)))$}
    \ELSIF{integrity attack}
        \IF{targeted attack}
            \STATE{$\mathcal{F}(\cdot) = \text{levenshtein\_distance}(\mathbf{y}, \mathcal{T}(\mathcal{A}(\mathbf{x}, \cdot)))$}
        \ELSE
            \STATE{$\mathcal{F}(\cdot) = \text{levenshtein\_distance}(\mathcal{T}(\mathbf{x}), \mathcal{T}(\mathcal{A}(\mathbf{x}, \cdot)))$}
        \ENDIF
    \ENDIF

    \FOR[\hspace{5em}$\rhd\ \mathcal{U}$ is uniform dist.]{$i := 0$ {\bfseries to} $m$}\COMMENT{}
        \FOR{$j := 0$ {\bfseries to} $s$}
            \STATE{$\mathbf{p_a}, \mathbf{p_b}, \mathbf{p_c} \xleftarrow{\text{rand}} \mathbf{P}$ s.t. $j\neq a \neq b \neq c$}
            \STATE $R \sim \mathcal{U}(0,|\mathbf{p_j}|)$
            \STATE $\mathbf{\hat{p}_j} := \mathbf{p_j}$
            \FOR{$k := 0$ \textbf{to} $|\mathbf{p_j}|$}
                \STATE $r_j \sim \mathcal{U}(0,1)$
                \IF{$r_j < \mathit{CR}$ \textbf{or} $R = k$}
                    \STATE $\mathbf{\hat{p}_{j_k}} = \mathbf{p_{a_k}} + F \times (\mathbf{p_{b_k}} - \mathbf{p_{b_k}})$
                \ENDIF
            \ENDFOR
            \IF{$\mathcal{F}(\mathbf{\hat{p}_j}) \ge \mathcal{F}(\mathbf{p_j})$}
                \STATE $\mathbf{p_j} = \mathbf{\hat{p}_j}$
            \ENDIF
        \ENDFOR
    \ENDFOR
    \STATE $\mathbf{\bar{f}} := \{ \mathcal{F}(\mathbf{p_0}), \dots, \mathcal{F}(\mathbf{p_s}) \}$
    \RETURN $\mathcal{A}(\mathbf{x}, \mathbf{p}_{\text{argmax}(\mathbf{\bar{f}})})$
            
\end{algorithmic}
\end{algorithm}

%% file: figures/infos/homoglyphs.tex
\begin{figure}[t]
    \centering
    \includegraphics[width=\linewidth]{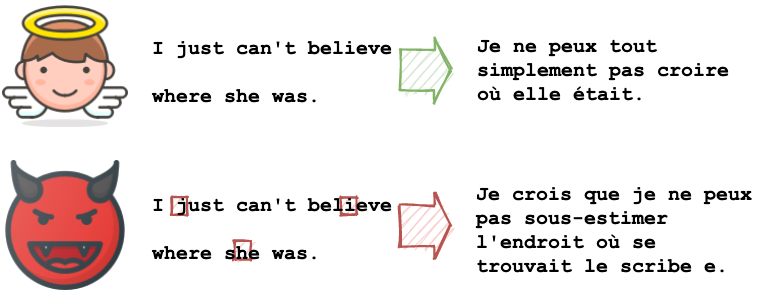}
    \caption{Attack using homoglyphs.  Example machine translation input is on the left with model output on the right. Homoglyphs are highlighted with red boxes, where \textit{j} is replaced with U+3F3, \textit{i} with U+456 and \textit{h} with U+4BB.}
    \label{fig:homoglyphs_attack}
\end{figure}

%% file: figures/infos/reordering.tex
\begin{figure}[t]
    \centering
    \includegraphics[width=\linewidth]{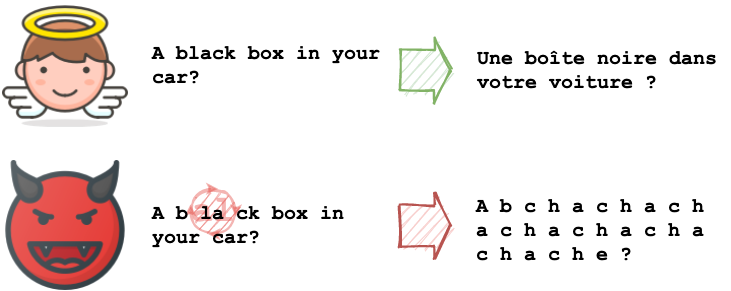}
    \caption{Attack using reorderings. Example machine translation input is on the left with model output on the right. The red circle denotes the string is encoded in reverse order surrounded by Bidi override characters.}
    \label{fig:reorder_attack}
\end{figure}

%% file: figures/homoglyph_clusters.tex
\begin{figure}[b]
    \centering
    \includegraphics[width=\linewidth]{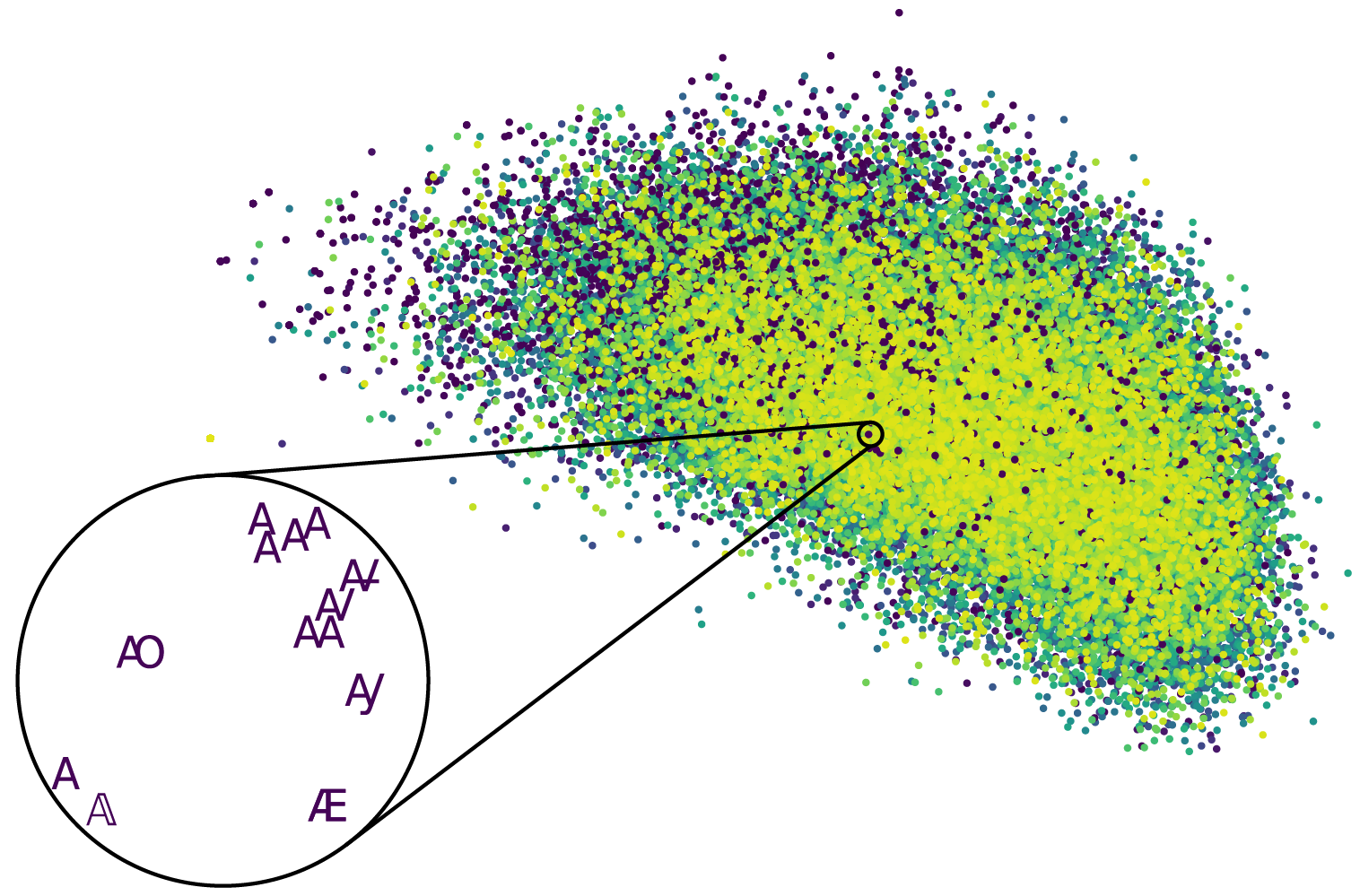}
    \caption{Clustering of Unicode homoglyphs according to the Unicode Security Confusables document, plotted as a 2D PCA of Unifont glyph images via a VGG16 model.}
    \label{fig:homoglyph-pca}
\end{figure}

%% file: figures/infos/backspace.tex
\begin{figure}[t]
    \centering
    \includegraphics[width=\linewidth]{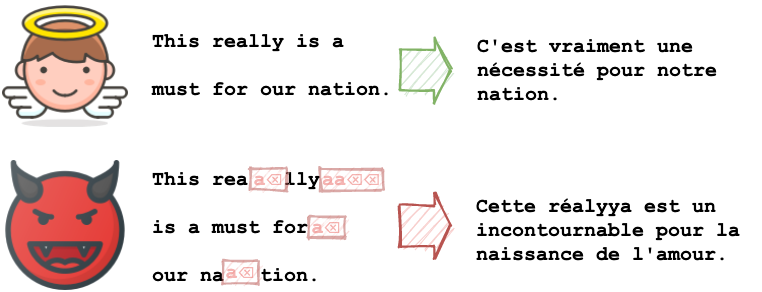}
    \caption{Attack using deletions. Example  machine  translation  input  is  on the  left  with  model  output on  the  right. The red boxes highlight injected characters followed by backspace characters.}
    \label{fig:backspace_attack}
\end{figure}

%% file: sections/attacks.tex
\section{NLP Attacks}

\input{figures/fairseq_translation}

\subsection{Integrity Attack}

Regardless of the tokenizer or dictionary used in an NLP model, systems are unlikely to handle imperceptible perturbations gracefully in the absence of specific defenses. Integrity attacks against NLP models exploit this fact to achieve degraded model performance in either a targeted or untargeted fashion.


The specific affect on input embedding transformation depends on the class of perturbation used:
\begin{itemize}
    \item \textbf{Invisible characters} (between words): Invisible characters are transformed into \unk embeddings between properly-embedded adjacent words.
    \item \textbf{Invisible characters} (within words): In addition to being transformed into \unk embeddings, the invisible characters may cause the word in which it is contained to be embedded as multiple shorter words, interfering with the standard processing.
    \item \textbf{Homoglyphs}: If the token containing the homoglyph is present in the model's dictionary, a word that contains it will be embedded with the less-common, and likely lower-performing, vector created from such data. If the homoglyph is not known, the token will be embedded as \unk.
    \item \textbf{Reorderings}: In addition to the Bidi control characters each being treated as invisible characters, the other characters input into the model will be in the underlying encoded order rather than the rendered order.
    \item \textbf{Deletions}: In addition to deletion-control characters each being treated as an invisible character, the deleted characters encoded into the input are still validly processed by the model.
\end{itemize}

Each of these modifications to embedded inputs degrades a model's performance. The cause is model-specific, but for attention-based models we expect that tokens in a context of \unk tokens are treated differently.

\subsection{Availability Attack}

Machine-learning systems can be attacked by workloads that are unusually slow. The inputs generating such computations are known as sponge examples~\cite{shumailov2020sponge}. 

In this paper we show that sponge examples can be constructed in a targeted way, both with fixed and increased input size. For a fixed-size sponge example, an attacker can replace individual characters with homoglyphs that take longer to process. If an increase in input size is tolerable, the attacker can also inject invisible characters, forcing the model to take additional time to process these additional steps in its input sequence. 

Such attacks may be carried out more covertly if the visual appearance of the input does not arouse users' suspicions. If launched in parallel at scale, the availability of hosted NLP models may be degraded, suggesting that a distributed denial-of-service attack may be feasible on text-processing services.

%% file: figures/fairseq_translation.tex
\begin{figure*}[t!]
  \centering
  \begin{minipage}[b]{0.49\textwidth}
    \centering
    \includesvg[width=\textwidth]{BLEU_Fairseq.svg}
    \caption{BLEU scores of imperceptible perturbations vs. unperturbed WMT data on Fairseq EN-FR model}
    \label{fig:bleu_fairseq}
  \end{minipage}
  \hfill
  \begin{minipage}[b]{0.49\textwidth}
    \centering
    \includesvg[width=\linewidth]{sponges.svg}
    \caption{Fairseq sponge example average inference time}
    \label{fig:sponge}
  \end{minipage}
\end{figure*}

%% file: sections/evaluation.tex
\section{Evaluation}
\label{sec:eval}

\subsection{Experiment Setup}

We evaluate the performance of each class of imperceptible perturbation attack -- invisible characters, homoglyphs, reorderings, and deletions -- against five NLP tasks: machine translation, toxic content detection, textual entailment classification, named entity recognition, and sentiment analysis. We perform these evaluations against a collection of five open-source models and three closed-source, commercial models published by Google, Facebook, Microsoft, IBM, and HuggingFace. We repeat each experiment with perturbation budget values varying from zero to five.

All experiments were performed in a black-box setting in which unlimited model evaluations are permitted, but accessing the assessed model's weights or state is not permitted. This represents one of the strongest threat models for which attacks are possible in nearly all settings, including against commercial Machine-Learning-as-a-Service (MLaaS) offerings. Every model examined was vulnerable to imperceptible perturbation attacks. We believe that the applicability of these attacks should in theory generalize to any text-based NLP model without adequate defenses in place.

We perform a collection of untargeted, targeted, and sponge example attacks across the eight models. The experiments were performed on a cluster of machines each equipped with a Tesla P100 GPU and Intel Xeon Silver 4110 CPU running Ubuntu.

For each class of perturbation, we followed \Cref{alg:attackalg} and found that the optimization converged quickly, thus choosing a population size of 32 with a maximum of 10 iterations in the genetic algorithm. Increasing these parameters further would likely allow an attacker to find even more effective perturbations; i.e. our experimental results obtain a lower bound.

For the objective functions used in these experiments, invisible characters were chosen from a set including ZWSP, ZWNJ, and ZWJ\footnote{Unicode characters U+200B, U+200C, U+200D}; homoglyphs sets were chosen according to the relevant Unicode technical report~\cite{unicode_intentionals}; reorderings were chosen from the sets defined using \Cref{alg:reorderalg}; and deletions were chosen from the set of all non-control ASCII characters followed by a BKSP\footnote{Unicode character U+0008} character. We define the unit value of the perturbation budget as one injected invisible character, one homoglyph character replacement, one \texttt{Swap} sequence according to the reordering algorithm, or one ASCII-backspace deletion pair.

We have published a command-line tool written in Python to conduct these experiments as well as the entire set of adversarial examples resulting from these experiments.\footnote{\href{https://github.com/nickboucher/imperceptible}{github.com/nickboucher/imperceptible}} We have also published an online tool for validating whether text may contain imperceptible perturbations and for generating random imperceptible perturbations.\footnote{\href{https://imperceptible.ml}{imperceptible.ml}}

In the following sections, we describe each experiment in detail.

\input{figures/toxic}

\subsection{Machine Translation: Integrity}

For the machine translation task, we used an English-French transformer model pre-trained on WMT14 data~\cite{ott-etal-2018-scaling} published by Facebook as part of Fairseq~\cite{ott2019fairseq}, Facebook AI Research's open source ML toolkit for sequence modeling. We utilized the corresponding WMT14 test set data to provide reference translations for each adversarial example.

For the set of integrity attacks, we crafted adversarial examples for 500 sentences and repeated adversarial generation for perturbations budgets of 0 through 5. Each example took, on average, 432 seconds to generate.

For the adversarial examples generated, we compare the BLEU~\cite{papineni-etal-2002-bleu} scores of the resulting translation against the reference translation in \Cref{fig:bleu_fairseq}. We also provide the Levenshtein distances between these values in Appendix \Cref{fig:levenshtein_fairseq}, which increase approximately linearly with reorderings having the largest distance.


\subsection{Machine Translation: Availability}

In addition to attacks on machine-translation model integrity, we also explored whether we could launch availability attacks. These attacks take the form of sponge examples, which are adversarial examples crafted to maximize inference runtime.

We used the same configuration as in the integrity experiments, crafting adversarial examples for 500 sentences with perturbation budgets of 0 to 5. Each example took, on average, 420 seconds to generate.

Sponge-example results against the Fairseq English-French model are presented in \Cref{fig:sponge}, which shows that reordering attacks are by some ways the most effective. Levenshtein distances are also provided in Appendix \Cref{fig:levenshtein_fairseq_sponge}. Although the slowdown is not as significant as Shumailov et al. achieved by dropping Chinese characters into Russian text~\cite{shumailov2020sponge}, our attacks are semantically meaningful and will not be noticeable to human eyes.

\subsection{Machine Translation: MLaaS}

In addition to the integrity attacks on Fairseq's open-source translation model, we performed a series of case studies on two popular Machine Learning as a Service (MLaaS) offerings: Google Translate and Microsoft Azure ML. These experiments attest to the real-world applicability of these attacks. In this setting, translation inference involves a web-based API call rather than invoking a local function.

Due to the cost of these services, we crafted adversarial examples targeting integrity for 20 sentences of budgets from 0 to 5 with a reduced maximum evolution iteration value of 3.

The BLEU results of tests against Google Translate are in Appendix \Cref{fig:bleu_google} and against Microsoft Azure ML in Appendix \Cref{fig:bleu_azure}. The corresponding Levenshtein results can be found in Appendix \Cref{fig:levenshtein_google,fig:levenshtein_azure}.


Interestingly, the adversarial examples generated against each platform appeared to be meaningfully effective against the other. The BLEU scores of each service's adversarial examples tested against the other are plotted as dotted lines in Appendix \Cref{fig:bleu_google,fig:bleu_azure}. These results show that imperceptible adversarial examples can be transferred between models.

\subsection{Toxic Content Detection}
\label{sec:toxic}

In this task we attempt to defeat a toxic-content detector. For our experiments, we use the open-source Toxic Content Classifier model~\cite{ibm_toxic} published by IBM. In this setting, the adversary has access to the classification probabilities emitted by the model.

For this set of experiments, we craft adversarial examples for 250 sentences labeled as toxic in the Wikipedia Detox Dataset~\cite{thain_dixon_wulczyn_2017} with perturbation budgets from 0 to 5. Each example took, on average, 18 seconds to generate.

IBM Toxic Content Classification perturbation results can be seen in \Cref{fig:maxtoxic}. Homoglyphs, reorderings, and deletions effectively degrade model performance by up to 75\%, but, interestingly, invisible characters do not have an effect on model performance. This could be because invisible characters were present in the training data and learned accordingly, or, more likely, the model uses a tokenizer which disregards the invisible characters we used.

\subsection{Toxic Content Detection: MLaaS}

We repeated the toxic content experiments against Google's Perspective API~\cite{perspective_api}, which is deployed at scale in the real world for toxic content detection. We used the same experiment setting as in the IBM Toxic Content Classification experiments, except that we generated adversarial examples for 50 sentences. The results can be seen in \Cref{fig:perspective}.

\subsection{Textual Entailment: Untargeted}

\input{figures/fairseq_mnli}

Recognizing textual entailment is a text-sequence classification task that requires labeling the relationship between a pair of sentences as entailment, contradiction, or neutral.

For the textual-entailment classification task, we performed experiments using the pre-trained RoBERTa model \cite{DBLP:journals/corr/abs-1907-11692} fine-tuned on the MNLI corpus \cite{N18-1101}. This model is published by Facebook as part of Fairseq~\cite{ott2019fairseq}.

For these textual-entailment integrity attacks, we crafted adversarial examples for 500 sentences and repeated adversarial generation for perturbation budgets of 0 through 5. The sentences used in this experiment were taken from the MNLI test set. Each example took, on average, 51 seconds to generate.

The results from this experiment are shown in \Cref{fig:mnli}. Performance drops significantly even with a budget of 1.

\subsection{Textual Entailment: Targeted}
\label{sec:targeted_mnli}

We repeated the set of textual-entailment classification integrity experiments with targeted attacks. For each sentence, we attempted to craft an adversarial example targeting each of the three possible output classes. Naturally, one of these classes is the correct unperturbed class, and as such we expect the budget $= 0$ results to be approximately 33\% successful.

Due to the increased number of adversarial examples per sentence, we crafted adversarial examples for 100 sentences and repeated adversarial generation for perturbation budgets of 0 through 5.

The results can be seen in \Cref{fig:mnli-targeted}. These attacks were up to 80.0\% successful with a budget of 5.

In the first set of targeted textual entailment experiments, we permitted the adversary to access the full set of logits output by the classification model. In other words, the differential evolution algorithm had access to the probability value assigned to each possible output class. We repeated the targeted textual entailment experiments a second time in which the adversary had access to the selected output label only, without probability values. These results are plotted as a dotted line in \Cref{fig:mnli-targeted}, and were up to 79.6\% successful with a budget of 5. Label-only attacks appear to suffer only a slight disadvantage, and even this diminishes as perturbation budgets increase.

\subsection{Named Entity Recognition: Targeted}

In addition to the Textual Entailment experiments, we also ran targeted attack experiments against the Named Entity Recognition (NER) task. We used a BERT~\cite{devlin2018bert} model~\cite{bert-conll} fine-tuned on the CoNLL-2003 dataset~\cite{10.3115/1119176.1119195}, which at the time of writing was the default NER model on HuggingFace~\cite{wolf-etal-2020-transformers}. We defined our attack as successful if one or more of the output tokens was classified as the target label, due to the fact that imperceptible perturbations typically break tokenizers and thus result in variable-length perturbed NER model outputs. We used the first 500 entries of the CoNLL-2003 test data split targeting each of the four possible labels using the same attack parameters as the prior experiments.

The attacks were up to 90.2\% successful with a budget of 5 depending on the technique selected, although invisible characters had no effect on this model.

The results are visualized in Appendix \Cref{fig:ner_targeted}.

\subsection{Sentiment Analysis: Targeted}

In addition to Textual Entailment and NER, we also ran targeted attack experiments against the sentiment analysis task. We used a DistilBERT~\cite{sanh2020distilbert} model~\cite{bert-emotion} fine-tuned on the Emotion dataset~\cite{saravia-etal-2018-carer} published on HuggingFace~\cite{wolf-etal-2020-transformers}. We used the first 500 entries of the test data split of the Emotion dataset targeting each of the six possible labels using the same attack parameters as the prior experiments.

The attacks were up to 79.2\% successful with a budget of 5 depending on the technique selection, although invisible characters also had no effect on this model.

The results are visualized in Appendix \Cref{fig:emotion_targeted}.

\subsection{Comparison with Previous Work}
\label{sec:prev_work_comparison}

We selected five attack methods described in prior adversarial NLP work to compare with imperceptible perturbations. Of immediate note is that all prior work results in visually perceptible perturbations whereas imperceptible perturbations have no visual artifacts.

Despite this, we leveraged tooling provided by TextAttack~\cite{morris2020textattack} to compare all four classes of imperceptible perturbations against TextBugger~\cite{li2019textbugger}, DeepWordBug~\cite{JiDeepWordBug18}, Probability Weighted Word Saliency~\cite{ren-etal-2019-generating}, Natural Language Adversarial Examples~\cite{alzantot-etal-2018-generating}, and an optimized version of Natural Language Adversarial Examples~\cite{jia2019certified}.

The results, shown in \Cref{fig:mnli_comparison}, indicate that with a budget of 10, imperceptible perturbations have similar adversarial efficacy as the existing perceptible methods. Moreover, the imperceptible budget could be arbitrarily increased without visual effect for even better adversarial performance.

\input{figures/mnli_comparison}

%% file: figures/toxic.tex
\begin{figure*}[t!]
  \centering
  \begin{minipage}[b]{0.49\textwidth}
    \centering
    \includesvg[width=\textwidth]{maxtoxic.svg}
    \caption{Percentage of imperceptibly perturbed toxic sentences classified correctly in IBM's Toxic Content Classifier.}
    \label{fig:maxtoxic}
  \end{minipage}
  \hfill
  \begin{minipage}[b]{0.49\textwidth}
    \centering
    \includesvg[width=\linewidth]{perspective.svg}
    \caption{Percentage of imperceptibly perturbed toxic sentences classified correctly in Google's Perspective API.}
    \label{fig:perspective}
  \end{minipage}
\end{figure*}

%% file: figures/fairseq_mnli.tex
\begin{figure*}[t!]
  \centering
  \begin{minipage}[b]{0.49\textwidth}
    \centering
    \includesvg[width=\linewidth]{mnli_untargeted.svg}
    \caption{Untargeted accuracy of Fairseq MNLI model with imperceptible perturbations}
    \label{fig:mnli}
  \end{minipage}
  \hfill
  \begin{minipage}[b]{0.49\textwidth}
    \centering
    \includesvg[width=\linewidth]{mnli_targeted.svg}
    \caption{Targeted accuracy of Fairseq MNLI model with imperceptible perturbations}
    \label{fig:mnli-targeted}
  \end{minipage}
\end{figure*}

%% file: figures/mnli_comparison.tex
\begin{figure}[t]
    \centering
    \includesvg[width=0.85\linewidth]{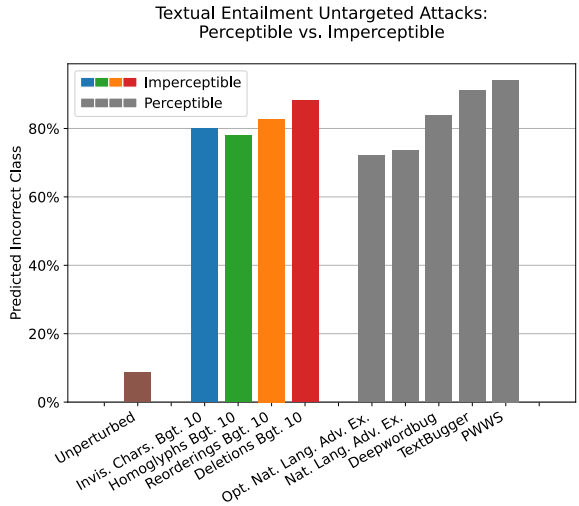}
    \caption{Perceptible and imperceptible attack success rates against Facebook Fairseq RoBERTa MNLI.}
    \label{fig:mnli_comparison}
\end{figure}

%% file: sections/discussion.tex
\section{Discussion}

\subsection{Ethics}

We followed departmental ethics guidelines closely. We used legitimate, well-formed API calls to all third parties, and paid for commercial products. To minimize the impact both on commercial services and CO$_2$ production, we chose small inputs, maximum iterations, and pool sizes. For example, while Microsoft Azure allows inputs of size 10,000~\cite{azure_limits}, we used inputs of less than 50 characters. Finally, we followed standard responsible disclosure processes.

\subsection{Experimental Interpretation}
\label{sec:interpretation}

Applying imperceptible perturbations drastically degrades the performance of all models examined, representing NLP tasks including machine translation, textual entailment classification, toxic content detection, named entity recognition, and sentiment analysis. Every performance metric, whether BLEU translation score, percentage correct classification, or average inference time, was degraded relative to no perturbations (budget=0), with degradation growing as the perturbation budget was increased.

The only exception was invisible character attacks against toxic content, NER, and sentiment analysis models, which had no effect; this is likely indicative of invisible characters being present in training data, or the tokenizer for these models ignoring the chosen invisible characters. For every other technique/model combination, however, there is a clear relationship between increased imperceptible perturbations and decreased model performance.

To make translation quality loss more concrete, we provide an example of varying BLEU scores in \Cref{tab:bleus}.

\subsection{Search Engine Attack}
\label{sec:search_engine}

Discrepancies between encoded bytes and their visual rendering affect searching and indexing systems. Search engine attacks fall into two categories: attacks on searching and attacks on indexing.

Attacks on searching result from perturbed search queries. Most systems search by comparing the encoded search query against indexed sets of resources. In an attack on searching, the adversary's goal is to degrade the quality or quantity of results. Perturbed queries interfere with the comparisons.

Attacks on indexing use perturbations to hide information from search engines. Even though a perturbed document may be crawled by a search engine's crawler, the terms used to index it will be affected by the perturbations, making it less likely to appear from a search on unperturbed terms. It is thus possible to hide documents from search engines ``in plain sight.'' As an example application, a dishonest company could mask negative information in its financial filings so that the specialist search engines used by stock analysts fail to pick it up.



\subsection{Attack Potential}
\label{sec:potential}

Imperceptible perturbations derived from manipulating Unicode encodings provide a broad and powerful class of attacks on text-based NLP models. They enable adversaries to:
\begin{itemize}
    \item Alter the output of machine translation systems;
    \item Evade toxic-content detection;
    \item Invisibly poison NLP training sets;
    \item Hide documents from indexing systems;
    \item Degrade the quality of search;
    \item Conduct denial-of-service attacks on NLP systems.
\end{itemize}


Perhaps the most disturbing aspect of our imperceptible perturbation attacks is their broad applicability: all text-based NLP systems we tested are susceptible. Indeed, any machine learning model which ingests user-supplied text as input is theoretically vulnerable to this attack. 
The adversarial implications may vary from one application to another and from one model to another, but all text-based models are based on encoded text, and all text is subject to adversarial encoding unless the coding is suitably constrained.

\subsection{Defenses}
\label{sec:defenses}

\input{figures/ocr_fairseq}


Given that the conceptual source of this attack stems from differences in logical and visual text encoding representation, one catch-all defense is to render all input, interpret it with optical character recognition (OCR), and feed the output into the original text model. This technique is described more formally in Appendix \Cref{alg:ocr_defense}. Such a tactic functionally forces models to operate on visual input rather than highly variable encodings, and has the added benefit that it can be retrofitted onto existing models without retraining.

To evaluate OCR as a general defense against imperceptible perturbations, we reevaluated the 500 adversarial examples previously generated for each technique against the Fairseq En$\rightarrow$FR translation model. Prior to inference, we preprocessed each sample by resolving control sequences in Python, rendering each input as an image with Pillow~\cite{alex_clark_pillow_2021} and Unifont~\cite{unifont}, and then performing OCR on each image with Tesseract~\cite{TessOverview} fine-tuned on Unifont. The results, shown in \Cref{fig:ocr_fairseq}, indicate that this technique fully prevents 100\% of invisible character, reordering, and deletion attacks while strongly mitigating the majority of homoglyph attacks.

Our experimental defense, however, comes at a cost of 6.2\% lowered baseline BLEU scores. This can be attributed to the OCR engine being imperfect; on some occasions, it outputs incorrect text for an unperturbed rendering. Similarly, it misinterprets homoglyphs at a higher rate than unperturbed text, leading to degraded defenses with the increased use of homoglyphs. Despite these shortcomings, OCR provides strong general defense at a relatively low cost without retraining existing models. Further, this cost could be decreased with better performing OCR models.

The accuracy and computational costs of retrofitting existing models with OCR may not be acceptable in all applications. We therefore explore additional defenses that may be more appropriate for certain settings.

\subsubsection{Invisible Character Defenses}

Generally speaking, invisible characters do not affect the semantic meaning of text, but relate to formatting concerns. For many text-based NLP applications, removing a standard set of invisible characters from inference inputs would block invisible character attacks. 

If application requirements do not allow discarding such characters, tokenizers must include them in the source-language dictionary to create non-\unk embeddings.

\subsubsection{Homoglyph Defenses}

Homoglyphs are perhaps the most challenging technique against which to defend. Functionally speaking, the OCR defense attempts to map unusual homoglyphs to their more common counterparts, thus increasing the likelihood that they are present in the NLP model's dictionary.

This mapping could be specified by model designers; a well-designed mapping of less-common homoglyphs to their most common counterparts applied prior to inference would have a similar effect to a high-performing OCR model. However, creating such a mapping is a daunting task, as the Unicode specification is immense. Automated techniques, such as previously depicted in \Cref{fig:homoglyph-pca}, may help to create these mappings.

\subsubsection{Reordering Defenses}



For some text-based NLP models with a graphical user interface, reordering attacks can be prevented by stripping all Bidi control characters as the input is displayed to the active user. In other settings, it may be more suitable to throw a warning for Bidi control characters.  

A more general solution, however -- and one that works for applications without a graphical user interface -- is to apply the Bidi algorithm to resolve Bidi control characters and coerce the logical order of text to match the order in which it would be visually rendered.

\subsubsection{Deletion Defenses}

We suspect that there may not be many use cases where deletion characters are a valid input into a model. Deletion characters may be resolved prior to inference, or a warning may be fired upon their detection.




%% file: figures/ocr_fairseq.tex
\begin{figure}[t]
    \centering
    \includesvg[width=\linewidth]{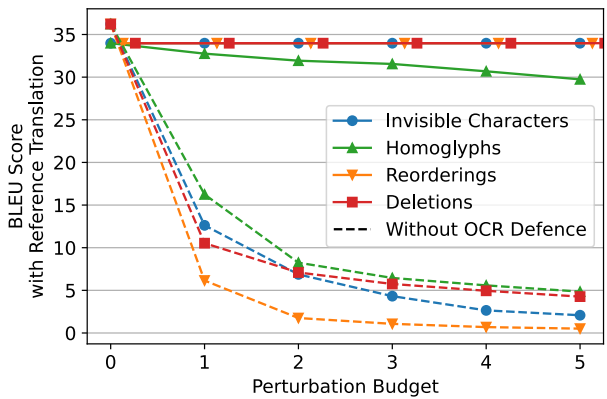}
    \caption{Evaluation of OCR defense against imperceptible perturbations.}
    \label{fig:ocr_fairseq}
\end{figure}

%% file: sections/conclusion.tex
\section{Conclusion}

Text-based NLP models are vulnerable to a broad class of imperceptible perturbations which can alter model output and increase inference runtime without modifying the visual appearance of the input. These attacks exploit language coding features, such as invisible characters and homoglyphs. Although they have been seen occasionally in the past in spam and phishing scams, the designers of the many NLP systems that are now being deployed at scale appear to have ignored them completely.

We have presented a systematic exploration of text-encoding exploits against NLP systems. We have developed a taxonomy of these attacks and explored in detail how they can be used to mislead and to poison machine-translation, toxic content detection, textual entailment classification, NER, and sentiment analysis systems. Indeed, they can be used on any text-based ML model that processes natural language. Furthermore, they can be used to degrade the quality of search engine results and hide data from indexing and filtering algorithms. 

We propose a variety of defenses against this class of attacks, and recommend that all firms building and deploying text-based NLP systems implement such defenses if they want their applications to be robust against malicious actors.

%% file: sections/appendix.tex
\section{Appendix}

\vspace{1cm}

\subsection{Machine Translation Fairseq Levenshtein Distances}

\begin{figure*}[h]
  \centering
  \begin{minipage}[b]{0.49\textwidth}
    \centering
    \includesvg[width=.85\textwidth]{Levenshtein_Fairseq.svg}
    \caption{Levenshtein distances between integrity attack imperceptible perturbations and unperturbed WMT data on Fairseq EN-FR model}
    \label{fig:levenshtein_fairseq}
  \end{minipage}
  \hfill
  \begin{minipage}[b]{0.49\textwidth}
    \centering
    \includesvg[width=.85\textwidth]{Levenshtein_Sponge.svg}
    \caption{Levenshtein distances between availability attack imperceptible perturbations and unperturbed WMT data on Fairseq EN-FR model}
    \label{fig:levenshtein_fairseq_sponge}
  \end{minipage}
\end{figure*}

\vspace{1cm}
\subsection{Example BLEU Scores}
\input{tables/bleu_table}

\clearpage

\subsection{Machine Translation MLaaS Results}
\input{figures/mlaas_translation}

\begin{figure*}[h]
  \centering
  \begin{minipage}{0.49\textwidth}
    \centering
    \includesvg[width=.85\linewidth]{Levenshtein_Google.svg}
    \caption{Levenshtein distances between imperceptible perturbations and unperturbed WMT data on Google Translate's EN-FR model}
    \label{fig:levenshtein_google}
  \end{minipage}
  \hfill
  \begin{minipage}{0.49\textwidth}
    \centering
    \includesvg[width=.85\linewidth]{Levenshtein_Azure.svg}
    \caption{Levenshtein distances between imperceptible perturbations and unperturbed WMT data on Microsoft Azure's EN-FR model}
    \label{fig:levenshtein_azure}
  \end{minipage}
\end{figure*}

\subsection{Multi-Class Targeted Classification Results}

\begin{figure*}[h]
  \centering
  \begin{minipage}[b]{0.49\textwidth}
    \centering
    \includesvg[width=.85\textwidth]{NER_Targeted.svg}
    \caption{Attack success rates for targeted Named Entity Recognition attacks against MDZ's CoNLL-2003 model with Imperceptible Perturbations}
    \label{fig:ner_targeted}
  \end{minipage}
  \hfill
  \begin{minipage}[b]{0.49\textwidth}
    \centering
    \includesvg[width=.85\textwidth]{Emotion_Targeted.svg}
    \caption{Attack success rates for targeted sentiment analysis Imperceptible Perturbations attacks against DistilBERT fine-tuned on the Emotion dataset}
    \label{fig:emotion_targeted}
  \end{minipage}
\end{figure*}

\clearpage
\twocolumn

\subsection{Bidirectional Reordering Algorithm}
\input{tables/reorder_algo}
\newpage

\subsection{OCR Defense Algorithm}
\input{tables/ocr_algo}
\vfill
\clearpage

%% file: tables/bleu_table.tex
\begin{table}[h]
\label{tab:bleus}
\caption{BLEU Scores across varying invisible character budgets for the input ``And I think about my father.'' with reference translation ``Et je pense à mon père.'' on the Fairseq WMT14 EN$\rightarrow$FR machine translation model.}
\resizebox{\textwidth}{!}{%
\begin{tabular}{@{}llll@{}}
\toprule
Budget & BLEU Score & Adversarial Example                                        & Adversarial Translation                                                                                                                \\ \midrule
0      & 100        & And I think about my father.                               & Et je pense à mon père.                                                                                                                \\
1      & 19.3       & And I think a{\color{red}{U+200D}}bout my father.                         & Et je pense que c\textquotesingle~est un bout de course pour mon père.                                                                                 \\
2      & 12.4       & And I think{\color{red}{U+200D}} about my fat{\color{red}{U+200B}}her.                   & Et je pense que l\textquotesingle~inquiétude au sujet de ma masse adipeuse l\textquotesingle~inquiète.                                                                 \\
3      & 1.9        & An{\color{red}{U+200B}}d I thi{\color{red}{U+200C}}nk about my f{\color{red}{U+200B}}ather.             & L \textquotedbl~âme d\textquotedbl~ une personne ne doit pas être confondue avec l \textquotedbl~âme d\textbackslash\textquotesingle~une autre personne.                                                   \\
4      & 1.9        & An{\color{red}{U+200D}}d{\color{red}{U+200C}} I think a{\color{red}{U+200B}}bout {\color{red}{U+200B}}my father.       & Un parent parent parent parent Je pense qu\textquotesingle~un parent parent parent parent parent parent parent parent parent parent parent parent      \\
5      & 0.9        & And{\color{red}{U+200D}} I thi{\color{red}{U+200C}}nk{\color{red}{U+200B}} {\color{red}{U+200B}}about my fathe{\color{red}{U+200B}}r. & Et Et Et Et Et Et Et Et Et Et Et Et Et Et Et Et Et Et Et Et Et Et Et Et Et Et Et Et Et Et Et Et Et Et Et Et Et Et Et Et Et Ma r. r. r. \\ \bottomrule
\end{tabular}
}
\end{table}

%% file: figures/mlaas_translation.tex
\begin{figure*}[h]
  \centering
  \begin{minipage}[b]{0.49\textwidth}
    \centering
    \includesvg[width=0.85\linewidth]{BLEU_Google_using_Azure.svg}
    \caption{BLEU Scores of Azure's imperceptible adversarial examples on Google Translate}
    \label{fig:bleu_google}
  \end{minipage}
  \hfill
  \begin{minipage}[b]{0.49\textwidth}
    \centering
    \includesvg[width=0.85\linewidth]{BLEU_Azure_using_Google.svg}
    \caption{BLEU Scores of Google Translate's imperceptible adversarial examples on Microsoft Azure}
    \label{fig:bleu_azure}
  \end{minipage}
\end{figure*}

%% file: tables/reorder_algo.tex
\begin{algorithm}[h]
\caption{Generation of $2^{n-1}$ visually identical strings via Unicode reorderings.}
\label{alg:reorderalg}
\begin{algorithmic}
    \STATE \textbf{Input:} \texttt{string} $x$ of length $n$
    \STATE \textbf{Result:} \texttt{Set} of $2^{n-1}$ visually identical reorderings of $x$
    \STATE
    \STATE \texttt{struct} \{ \texttt{string} one, two; \} \texttt{Swap}
    \STATE{\begin{tabular}{@{}l@{}l@{}l@{}l@{}}
        \texttt{string} PDF &\ := 0x202C, &\ LRO &\ := 0x202D \\
        \texttt{string} RLO &\ := 0x202E, &\ PDI &\ := 0x2069 \\
        \texttt{string} LRI &\ := 0x2066 && \\ \\
    \end{tabular}}

    \STATE {\bfseries procedure} \textsc{swaps} (\textit{body, prefix, suffix})
        \STATE\quad \texttt{Set} orderings := \{ concatenate(prefix, body, suffix) \}
        \STATE\quad \textbf{for} $i:=0$ \textbf{to} length(body)-1 \textbf{do}
           \STATE\quad\quad \texttt{Swap} swap := \{ body[i+1], body[i] \}
           \STATE\quad\quad{\begin{tabular}{@{}l@{}l@{}}
                orderings.add(&[prefix, body[:i], \\
                &swap, body[i+1:], suffix])
           \end{tabular}}
           \STATE\quad\quad orderings.union(\textsc{swaps}(suffix, [prefix, swap], \texttt{null}))
           \STATE\quad\quad orderings.union(\textsc{swaps}([prefix, swap], \texttt{null}, suffix))
        \STATE\quad {\bfseries end for}
    \STATE\quad\textbf{return} orderings
\STATE {\bfseries end procedure}
\STATE

\STATE {\bfseries procedure} \textsc{encode} (\textit{ordering})
    \STATE\quad \texttt{string} encoding := ""
    \STATE\quad\textbf{for} element \textbf{in} ordering \textbf{do}
        \STATE\quad\quad\textbf{if} element is \texttt{Swap}
            \STATE\quad\quad\quad{\begin{tabular}{@{}l@{}l@{}}
                swap = \textsc{encode}(&[LRO, LRI, RLO, LRI, \\
                                       &element.one, PDI, LRI, \\
                                       &element.two, PDI, PDF, \\
                                       &PDI, PDF]) \\
            \end{tabular}}
            \STATE\quad\quad\quad encoding = concatenate(encoding, swap)
        \STATE\quad\quad\textbf{else if} element is \texttt{string}
            \STATE\quad\quad\quad encoding = concatenate(encoding, element)
    \STATE\quad {\bfseries end for}
    \STATE\quad\textbf{return} encoding
\STATE {\bfseries end procedure}
\STATE

\STATE\texttt{Set} orderings := \{ \}
\FOR{ordering \textbf{in} \textsc{swaps}($x$, null, null)}
    \STATE orderings.add(\textsc{encode}(ordering))
\ENDFOR
\RETURN orderings
\end{algorithmic}
\end{algorithm}

%% file: tables/ocr_algo.tex
\begin{algorithm}[h]
\caption{OCR defense technique against imperceptible perturbations via input pre-processing.}
\label{alg:ocr_defense}
\begin{algorithmic}
    \STATE {\bfseries Input:} model input text $\mathbf{x}$
    \STATE \textbf{Result:} pre-processed model input text $\mathbf{x'}$
    \STATE
    
    \STATE $x =$ resolve\_control\_chars($x$)\hspace{1em}$\rhd$ Apply Bidi+Deletion \COMMENT{}
    \STATE $i :=$ render\_text($x$)
    \STATE $x' :=$ ocr($i$)
    \RETURN $x'$\hspace{8.8em}$\rhd$ Pass output to model \COMMENT{}
\end{algorithmic}
\end{algorithm}